\documentclass[10pt,twocolumn,letterpaper]{article}
\usepackage{geometry}
\usepackage{cvpr}
\usepackage{times}
\usepackage{epsfig}
\usepackage{graphicx}
\usepackage{amsmath}
\usepackage{amssymb}

\usepackage[linesnumbered, ruled]{algorithm2e}
\usepackage{algorithmic}
\usepackage[utf8]{inputenc} % allow utf-8 input
\usepackage[T1]{fontenc}    % use 8-bit T1 fonts
\usepackage{url}            % simple URL typesetting
\usepackage{booktabs}       % professional-quality tables
\usepackage{amsfonts}       % blackboard math symbols
\usepackage{nicefrac}       % compact symbols for 1/2, etc.
\usepackage{microtype}      % microtypography
\usepackage{amsmath,amssymb}
\usepackage{color}
\usepackage{float}
\usepackage{array}
\usepackage{multirow}
\usepackage{booktabs}
\usepackage{subfigure}
\usepackage{amsthm}
\usepackage{wrapfig,lipsum,booktabs}
\usepackage{xcolor}
\usepackage{bm}
\usepackage{textcomp}
\usepackage[noblocks]{authblk}
\newcommand{\todo}{\textcolor[rgb]{0.00,0.00,0.00}}

\newcommand{\whli}{\textcolor[rgb]{0.00,0.00,0.00}}
\newcommand{\whlie}{\textcolor[rgb]{0.00,0.00,0.00}}
\newcommand{\whliee}{\textcolor[rgb]{0.00,0.00,0.00}}
\newcommand{\whlieee}{\textcolor[rgb]{0.00,0.00,0.00}}
\newcommand{\jason}{\textcolor[rgb]{0.00,0.00,0.00}}
\newcommand{\whlia}{\textcolor[rgb]{0.00,0.00,0.00}}
\newcommand{\whlcam}{\textcolor[rgb]{0.00,0.00,0.00}}
\newcommand{\erhao}{\fontsize{21pt}{\baselineskip}\selectfont}

\usepackage[breaklinks=true,bookmarks=false, colorlinks, linkcolor=red]{hyperref}

\cvprfinalcopy % *** Uncomment this line for the final submission

\def\cvprPaperID{2305} % *** Enter the CVPR Paper ID here

\begin{document}
%\newgeometry{top=6cm,bottom=1cm}

{\onecolumn

\noindent \textbf{\erhao{Learning to Learn Relation for Important People Detection in Still Images}}

\vspace{2cm}

\noindent {\LARGE{Wei-Hong Li, Fa-Ting Hong, Wei-Shi Zheng}}

%\Large
\vspace{2cm}

\noindent Code is available at: \\
\ \ \ \ \ \ \ \ \ \ \ \ \url{https://weihonglee.github.io/Projects/POINT/POINT.htm}

\vspace{1cm}

\noindent For reference of this work, please cite:

\vspace{1cm}
\noindent Wei-Hong Li, Fa-Ting Hong and Wei-Shi Zheng.
``Learning to Learn Relation for Important People Detection in Still Images.''
In \emph{Proceedings of the IEEE International Conference on Computer Vision and Pattern Recognition.} 2019.

\vspace{1cm}

\noindent Bib:\\
\noindent
@inproceedings\{li2019point,\\
\ \ \   title=\{Learning to Learn Relation for Important People Detection in Still Images\},\\
\ \ \  author=\{Li, Wei-Hong, Hong, Fa-Ting and Zheng, Wei-Shi\},\\
\ \ \  booktitle=\{Proceedings of the IEEE International Conference on Computer Vision and Pattern Recognition\},\\
\ \ \  year=\{2019\}\\
\}
}

%\clearpage
%
%\newpage
\restoregeometry
%%%%%%%%% TITLE

%%%%%%%%% TITLE
\title{Learning to Learn Relation for Important People Detection \\ in Still Images}

\author[ ]{\vspace{-0.5cm}Wei-Hong Li$^{1,2}$\thanks{Equal contribution. Work done at Sun Yat-sen University.}}
\author[ ]{Fa-Ting Hong$^{1,3,4*}$}
\author[ ]{Wei-Shi Zheng$^{1,4}$\thanks{Corresponding author}\vspace{-0.3cm}}

%\affil[ ]{$^{1}$Intelligence Science and System Lab, Sun Yat-sen University, China}
% \affil[ ]{$^{1}$\small School of Electronics and Information Technology, Sun Yat-sen University, China}
\affil[ ]{\small$^{1}$ School of Data and Computer Science, Sun Yat-sen University, China}
\affil[ ]{\small$^{2}$ VICO Group, School of Informatics, University of Edinburgh, United Kingdom}
\affil[ ]{\small$^{3}$ Accuvision Technology Co. Ltd.}
\affil[ ]{\small$^{4}$ Key Laboratory of Machine Intelligence and Advanced Computing, Ministry of Education, China.}
\affil[ ]{\tt\small w.h.li@ed.ac.uk, hongft3@mail2.sysu.edu.cn, wszheng@ieee.org\vspace{-0.5cm}}

% \author{Wei-Hong Li\thanks{Equal contribution. Work done while at Sun Yat-sen University.},~~~ $\text{Fa-Ting Hong}$\footnotemark[1]~~~~and~~~Wei-Shi Zheng\thanks{Primary contact author.}\\
% % School of Informatics\\
% School of Data and Computer Science, Sun Yat-sen University, China.\\ School of Informatics, University of Edinburgh, United Kingdom.\\
% {\tt\small w.h.li@ed.ac.uk, hongft3@mail2.sysu.edu.cn, wszheng@ieee.org}
% % \and
% % $\text{Fa-Ting Hong}^{*}$\\
% % % School of Data and Computer Science\\
% % Sun Yat-sen University,\\
% % {\tt\small hongft3@mail2.sysu.edu.cn}
% % \and
% % Wei-Shi Zheng\\
% % % School of Data and Computer Science\\
% % Sun Yat-sen University,\\
% % {\tt\small wszheng@ieee.org}
% }

\maketitle
\thispagestyle{empty}

%%%%%%%%% ABSTRACT
\begin{abstract}
Humans can easily recognize the importance of people in social event images, and they always focus on the most important individuals. However, learning to learn the relation between people in an image, and inferring the most important person based on this relation, remains undeveloped. In this work, we propose a deep imPOrtance relatIon NeTwork (POINT) that combines both relation modeling and feature learning. \whlie{In particular, we infer two types of interaction modules: the person-person interaction module that learns the interaction between people and the event-person interaction module that learns to describe how a person is involved in the event occurring in an image. 
We then estimate the importance relations among people from both interactions and \whliee{encode the relation feature from the importance relations}. In this way, POINT automatically learns several types of relation features in parallel, and we aggregate these relation features and the person's feature to form the importance feature for important people classification.}
%Our framework enables learning deep features tightly coupled to important people detection and thus the learned effective deep importance features.
Extensive experimental results show that our method is effective for important people detection and verify the efficacy of learning to learn relations for important people detection.
%	And the result also verifies that inferring the importance of persons is better than inferring directly from people's individual feature.
%	to represent the relation between persons which can benefit important people detection and inferring importance of persons in images is better than inferring directly from persons' individual feautures.

%To facilitate the research, we have investigate the effect of various basic attention functions on modeling interactions and the effect of different sorts of information extracted from images on important people detection.
%	different sorts of information extracted from images on important people detection. 
\end{abstract}
%%%%%%%%% BODY TEXT

\section{Introduction}

\begin{figure}[t]
	\begin{center}
		\label{fig:ShowOnFirstPage}
		%		\fbox{\rule{0pt}{2in}\rule{0.9\linewidth}{0pt}}
		\includegraphics[width=0.95 \linewidth]{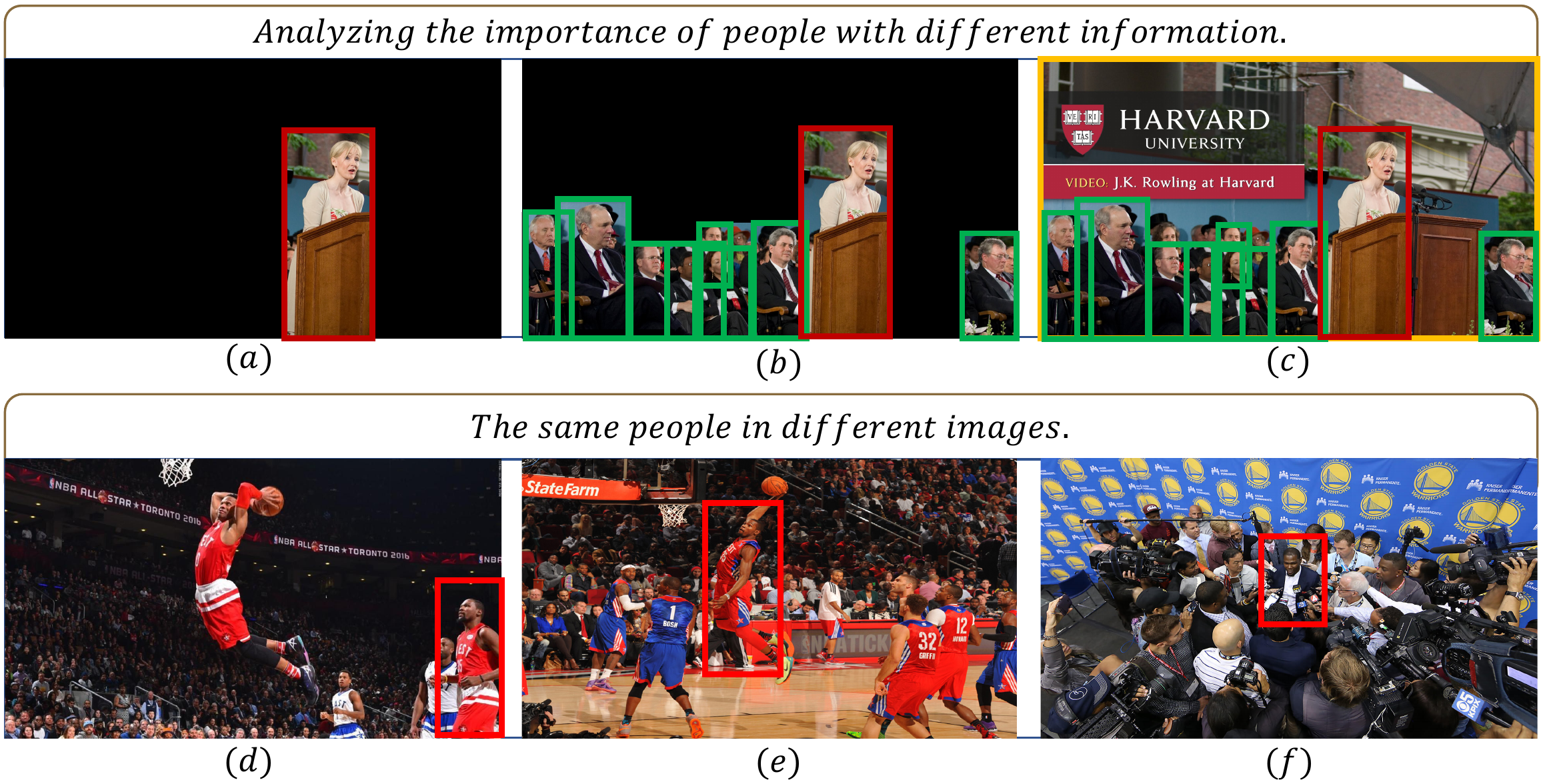}
		%		\vspace{-0.2cm}
		\centering\small\caption{Inferring the importance of persons from an image is inherently complex and difficult as it relates to diverse information (i.e., individual features of persons (Figure (a)), relations among persons (Figure (b)) and the event information (Figure(c)) from the whole image). The great visual variations lead to difficulties as well. The person in the red bounding box in all the images shown in the second row is the same person while he plays diverse roles in these images. He is the most important person in Figure (e) and (f) while his appearance, location and the event in both images are completely distinct. Comparing \whli{between} Figure (d) and Figure (e), he wears the same clothes in both images but his importance in these images is different.}
		\label{fig:ShowOnFirstPage}
	\end{center}
	\vspace{-0.3cm}
\end{figure}

% \vspace{-0.3cm}

In our daily lives, we often see wonderful live broadcasting as the cameraman can easily recognize the importance of people in an event and take shots or videos of the important people in the event to present what is occurring at the moment.
Additionally, when a social event image is presented, humans can easily recognize the distinct importance of different faces (persons) in the event and focus on the most important people (e.g., when people are watching a basketball game, they are more likely to focus on the shooter or the player with the basketball). 
It is natural to ask whether a computer vision model can be built to automatically detect the important people in event images.
It is also known that
correctly detecting the most important people in images can benefit other vision tasks such as event detection \cite{Key_ramanathan2015detecting}, event/activity recognition \cite{Key_ramanathan2015detecting,Acti_tang2017latent} and image captioning \cite{VIP_solomon2015vip}. 

%To achieve this, 
Important people detection has only recently become the focus of research. To detect the important people in still images, a straightforward approach is to 
%disregard the lack of relations among persons and 
exploit classification or regression models to infer the importance of people directly from their individual features \cite{VIP_solomon2015vip}. 
%However, it used the features trained for other tasks and focused on training a classification or regression model. It is shown that this feature is less effective for represents the importance of people.
Another solution considers the relations among persons by estimating their interaction sequentially (i.e., sequential relation models \cite{VIP_solomon2015vip,PR_li2018personrank}).  
Solomon et al. \cite{VIP_solomon2015vip} studied the relative importance between a pair of faces, either in the same image or separate images, and developed a regression model to predict the relative importance between any pair of faces using manually designed features. Li et al. \cite{PR_li2018personrank} modeled all previously detected people in a \whlie{hybrid interaction} graph and developed PersonRank, a graphical model to rank the people from the interaction graphs.

Despite these efforts on important people detection, the problem remains challenging, as the \textbf{importance} of people is related to not only their appearance but also, more importantly, the relations among the people.
%the contextual information, the event where people involve in,  and so on. 
Only relying on the appearance features is not effective.
For instance, we would be unable to determine whether the lady with the red bounding box in Figure \ref{fig:ShowOnFirstPage}(c) is important or not if we were given the patch inside the red bounding box as shown in Figure \ref{fig:ShowOnFirstPage}(a).
However, if we know who and how others individuals are interacting with the lady (Figure \ref{fig:ShowOnFirstPage}), it becomes easier to separate \whliee{the lady
% (the important people) 
from the others.}
% the collective activity from other similar activites. 
Although relation modeling is important, the relation between two people in an image is still determined by customized features \whli{(e.g., \cite{VIP_solomon2015vip,PR_li2018personrank})}. \whli{The} customized features are highly affected by variations in pose, appearance and actions. How to automatically exploit the reliable and effective \whlie{relation} features that describe the relations between people is still unsolved.

%The relation modelling is important for overcoming the challenge. 
%the same people in different event images would have different importance and diverse variations of pose, action, appearance of the important people, various changes of scene and heavy occlusions among people make it challenging to model the relations between people. 

In this work, we \jason{cast the important people detection problem as learning the relation network among detected people in an image and inferring the most active person there. Thus, we} attempt to develop a deep imPOrtance relatIon NeTwork (POINT) to allow machine learning to exploit the relations automatically. 
In POINT, we \whlie{mainly} introduce \whlie{the relation module, which contains several \whlie{relation submodules} to automatically construct interaction graphs and model their importance relations from the interaction graphs. \whlie{In each relation submodule}, we \whli{form} two types of interaction modules, the person-person \whliee{interaction} module and the \whliee{event-person interaction} module. The person-person \whliee{interaction} module describes the pairwise person interactions and the \whliee{event-person interaction} module indicates the probability of a person \whlie{being} involved in the event.}
% being an important person. 
\whli{We then introduce two methods to estimate the importance relations among persons from both the interaction graphs and \whlie{encode} the relation feature based on the importance relations. Finally, \whlie{we concatenate the relation features from all relation submodules into one relation feature and employ the residual
% After that, the relation feature is encoded by aggregating features from others multiplied by the relations/***Jason: what do you mean?***/. 
% we use the residual 
connection to aggregate the concatenated relation feature and the \whli{person feature}},
% person's original feature 
% and the relation feature, 
resulting in the \whlie{importance feature} for the final importance classification.} In summary, the POINT method is a classification framework consisting of a feature representation module, a relation module and an importance classification module. 

To the best of our knowledge, POINT is the first to investigate deep learning for exploring and encoding the relation features and exploiting them for important people detection.
In our experiments, we investigate and discuss the effect of various types of basic \whlie{interaction} functions (i.e., additive function and scaled dot product function) on modeling pairwise persons interactions and the effect of different types of information on important people detection.
The experimental results show that our deep relation network achieves state-of-the-art performance on two public datasets and verify its efficacy for important people detection.

\section{Related Work}

% \vspace{-0.1cm}

\noindent\textbf{Persons and General Object Importance.}
Recently, the importance of generic object categories and persons has attracted increased attention and has been studied by several researchers \cite{IP_berg2012understanding,IP_hwang2012learning,IP_le2007finding,IP_lee2012discovering,IP_spain2011measuring,IP_lee2015predicting,VIP_solomon2015vip,PR_li2018personrank}. Solomon et al. \cite{VIP_solomon2015vip} focused on studying the relative importance between a pair of faces, either in the same image or separate images, and developed a regression model for predicting the importance of faces. The authors designed customized features containing spatial and saliency information of faces for important face detection. In addition, Ramanathan et al. \cite{Key_ramanathan2015detecting} trained an attention-based model with event recognition labels to assign attention/importance scores to all detected individuals to measure how related they were to basketball game videos. \whlie{More specifically}, they proposed utilizing spatial and appearance features of persons including temporal information to infer the importance score of all detected persons.
Recently, Li et al. \cite{PR_li2018personrank} modeled all detected people in a \whlie{hybrid interaction} graph by organizing the interaction among persons sequentially and developed PersonRank, a graphical model to rank the persons by inferring the importance scores of persons from person-person interactions constructed on four types of features that have been pretrained for other tasks. 

Different from the aforementioned methods, which design both handcrafted relations as well as features, or those pretrained for other tasks, as far as we know, our work is the first to design a deep architecture to combine the learning of relations and features for important people detection. The relation module is learned to construct interaction graphs and automatically encode \whlie{relation features}. Thus, our network can not only encode more effective features from a person’s individual information but also efficiently encode the relations from other people and the event in the image.

%We also integrate an adapted relation module which can process a set of persons in parallel into our framework. 

% \vspace{0.05cm}
% \noindent \textbf{Attention Modelling.}
% Our proposed model is related to current Attention Modules \cite{AttenIsAll_vaswani2017attention,AttenFun_britz2017massive,AttenFunc_bahdanau2014neural} in natural language processing field. An attention module affects an individual a word in the target sentence in machine translation by aggregating information from all words in the source sentence.
% An apparent distinction is that the importance of people is a complex vision task and it is related to diverse sorts of information (e.g. pair-wise persons dependency, 2D location, event-person interaction and etc). In comparison with natural language processing, positional and global information plays an complex and important role on important people detection. We propose to integrate these additional information into the original person-person relation weight in different effective ways.
% Moreover, we introduce a new solution to estimate importance relation weight which yield convicing results. 

\begin{figure*}[htp]
	\begin{center}
		\label{fig:FrameWork}
		%		\fbox{\rule{0pt}{2in}\rule{0.9\linewidth}{0pt}}
		\includegraphics[width=0.85\linewidth]{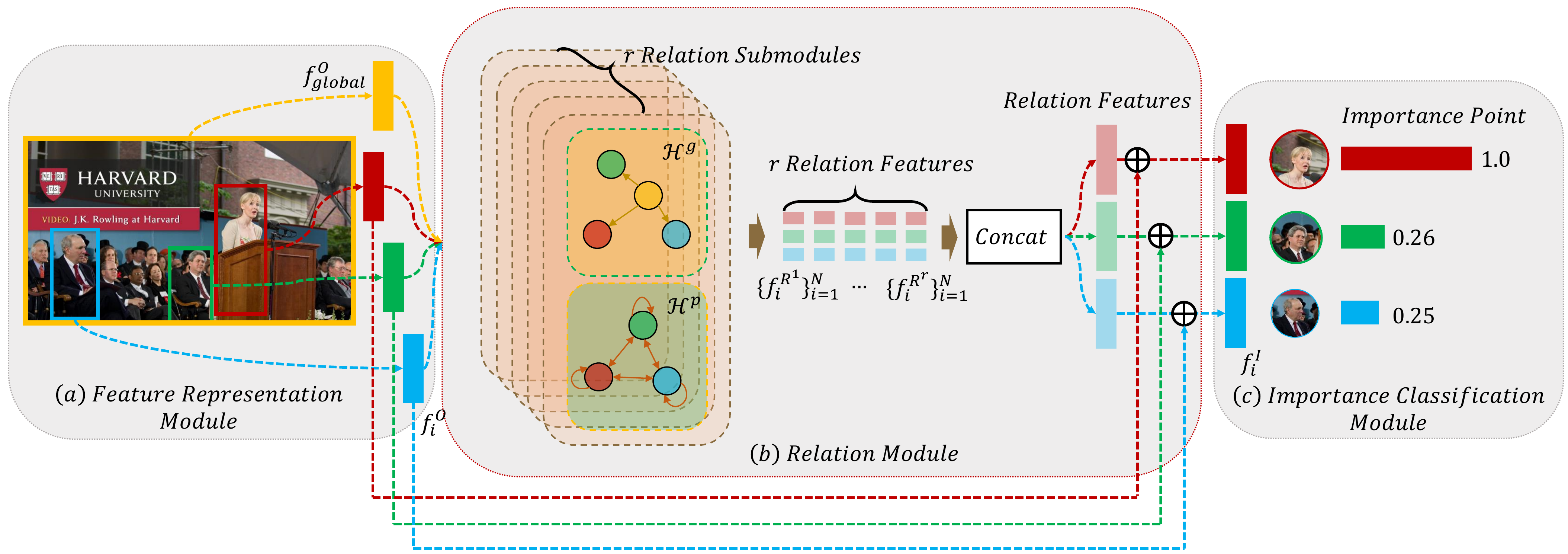}
		%		\vspace{-0.2cm}
		\centering\small\caption{\whlie{An illustration of our deep imPOrtance relatIons NeTworks (POINT). We exploit the feature representation module to extract the \whli{person feature} for persons and the global feature for the whole image (Figure (a)). These features are fed into the relation module, which contains $r$ relation submodules. In each relation submodule, we construct two interaction graphs and estimate importance relations from both graphs, which are used for encoding relation features. In this way, the POINT learns $r$ relation features in parallel and these features are concatenated into a relation feature vector. We add this concatenated relation feature to the person feature, resulting in the importance feature.
%         which contains $N_r$ relation module blocks and each module block consists of $r$ basic relation modules. 
%         In the relation module, we aggregate the relation features from other persons and the global/event and add this relation feature to the \whli{person feature}, resulting in the importance feature. 
        Finally, the importance classification module is employed to infer the importance point of people.}
		}
		%For each person, we extract features and formulated hierarchy interaction graph of these persons by message functions. Finally, we \weihong{inferred} the importance of persons (Better viewed in color) and the most important person (the person in red box) was selected.}
		\label{fig:FrameWork}
	\end{center}
	\vspace{-0.3cm}
\end{figure*}

\vspace{0.05cm}
\noindent \textbf{Relation Networks on Vision Tasks.}
Relation modeling is not limited to important people detection and has broad application, such as object detection \cite{Rela_hu2018relation}, AI gaming \cite{Rela_zambaldi2018relational}, image captioning \cite{Rela_xu2015show}, video classification \cite{RN_wang2017appearance}, and few-shot recognition \cite{RN_yang2018learning}. Related to our method, Hu et al. \cite{Rela_hu2018relation} proposed adapting the attention module by embedding a new geometric weight and applying it in a typical object detection CNN model to enhance the features for object classification and duplicate removal. Zambaldi et al. \cite{Rela_zambaldi2018relational} exploited the attention module to iteratively identify the relations between entities in a scene and to guide a model-free policy in a novel navigation and planning task called Box-World.

In this work, we have the different purpose of building a relation network for important people detection, while the related relation models are not suitable for our task. 
\whlie{In particular, in previous works, they learn the relation that describes the appearance and location similarity between two objects/entities to find the similar objects. These relational models will bias the important people detection model to detect the people with certain appearance or the people in a specific location,} \jason{but not for purpose of telling how people are interacting with each other and who is the most active one. In our experiments, we have shown only using appearance features or specific location is not effective for important people detection \whlieee{(see Table \ref{tab:MS_Methods}, the SVR-person only using appearance and location information is not effective.)}
% /***Jason: to enrich by weihong***/). 
For estimating the important relations, we introduce two interaction modules (i.e. person-person and event-person interactions) to learn the interactions that describe the relation between two people and how people are involved in the event occurring in an image automatically.} %and encode effective relation features from the importance relations.}

\section{Approach}
% \vspace{-0.1cm}
Detecting important people in still images is a more challenging task than conventional people detection as it requires extracting higher semantic information than other detection tasks. In this work, under the same setting as that in previous works \cite{Key_ramanathan2015detecting,VIP_solomon2015vip,PR_li2018personrank}\footnote{Similar to the aforementioned works \cite{Key_ramanathan2015detecting,VIP_solomon2015vip,PR_li2018personrank}, we assume that all persons appearing in images are successfully detected by existing state-of-the-art person (face or pedestrian) detectors. },
we aim to design a deep relation network called the deep imPOrtance relatIons NeTwork (POINT) (Section \ref{sec:OverV}), which learns to build the relations and combines the \whlie{relation modeling} with feature learning for important people detection. 
%Editor: Abbreviations and acronyms typically need to be defined only once within the main text. Please consider adhering to this convention.
%In addition to this, it should have capability of learning to model interaction graphs automatically and extracting features embracing relations to better represent persons' importance in an event shown. In this section, we are going to elaborate our Deep Relations Networks.
% which is able to learn deep feature embracing relations among persons and recognize the importance of persons from relations graphs. 
We briefly introduce the architecture of the proposed POINT (Section \ref{sec:OverV}) before detailing three specific modules and loss (Section \ref{sec:Feat}, Section \ref{sec:Relation} and Section \ref{sec:Class}). 
%our basic model (Section \ref{sec:Basic}) and relations module (Section \ref{sec:Relation})

\begin{figure}[t]
	\begin{center}
		\label{fig:Feat}
		%		\fbox{\rule{0pt}{2in}\rule{0.9\linewidth}{0pt}}
		\includegraphics[width=0.8\linewidth]{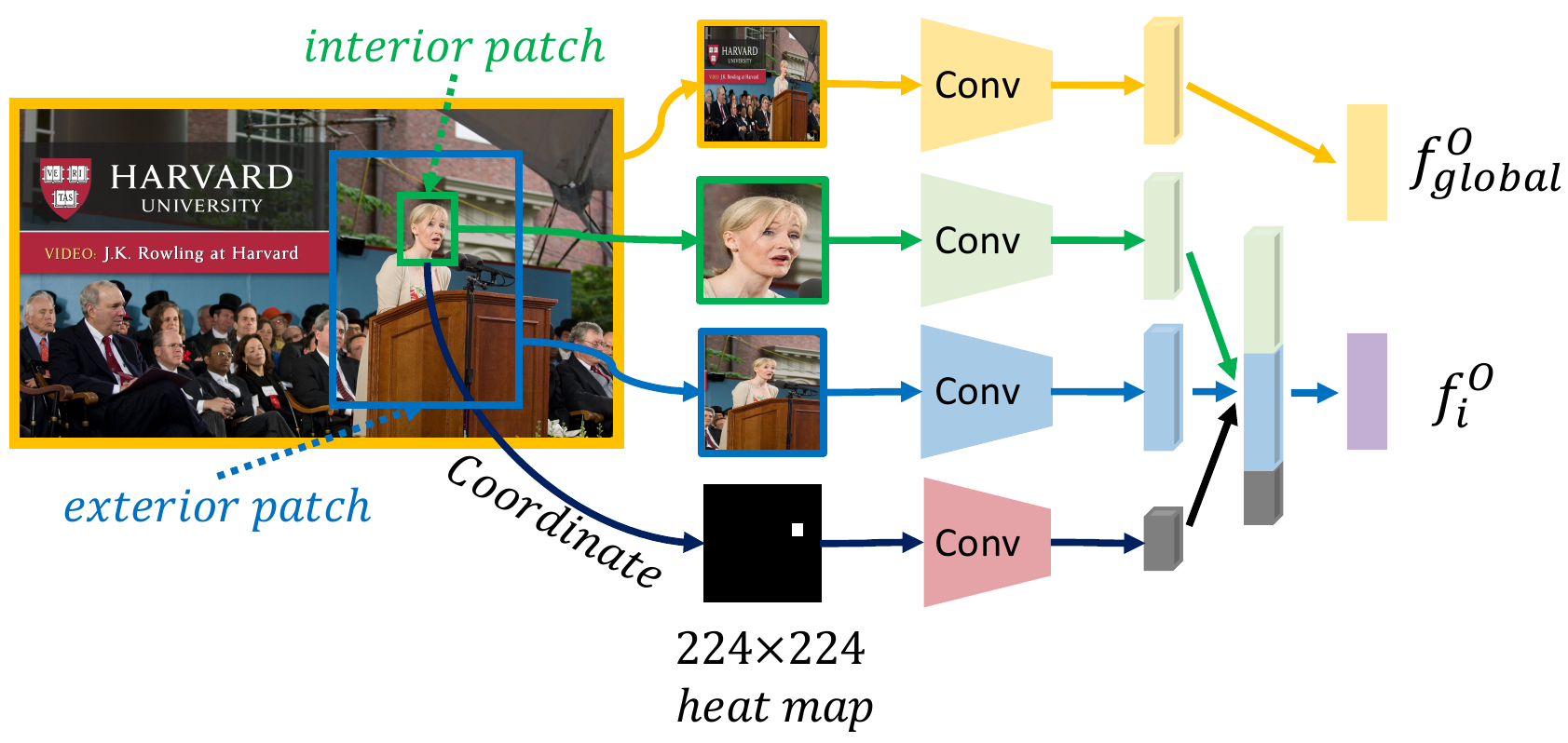}
		%		\vspace{-0.2cm}
		\centering\small\caption{The feature representation module.}
		%For each person, we extract features and formulated hierarchy interaction graph of these persons by message functions. Finally, we \weihong{inferred} the importance of persons (Better viewed in color) and the most important person (the person in red box) was selected.}
		\label{fig:Feat}
	\end{center}
	\vspace{-0.35cm}
\end{figure}

%At first, we briefly introduce the architecture of our proposed Deep Relations Networks which is able to learn deep feature which embraces relations among persons and recognize the importance of persons from relation graph (Section \ref{sec:OverV}).
%
%Since there is a lack of deep learning architecture which enables feature representation learning to be tightedly coupled to important people detection, we build up our basic model which treats this task as a classification problem. Different from models in other tasks using information inside bounding boxes. we exploit features embrace interior, exterior and location information to represent each persons (Section \ref{sec:Basic}). 
%
%
%
%first briefly introduce our baseline model that infers importance score of persons directly from individuals features (Section \ref{sec:OverV}).
%
%In this section, we are going to elaborate our Deep Relations Networks for important people detection. 
%work exploring on designing 
%In this work, we aim at developing deep learning architecture that is able to learn deep feature which embraces relations among persons and recognize the importance of persons from relation graph. 
%We briefly introduce the proposed deep relational framework for important people detection (Section \ref{sec:OverV}) and the use of different types of relation modules (Section \ref{sec:Relation}) before detailing the training procedure (Section \ref{sec:Train})

\subsection{Overview}\label{sec:OverV}
% \vspace{-0.1cm}

An illustration of our proposed model's architecture is shown in Figure \ref{fig:FrameWork}. Given a social event image $\mathbf{I}$ and all detected ($N$) persons $\{\mathbf{p}_i\}_{i=1}^{N}$, to analyze the importance of these persons, we build our \whliee{POINT}
% deep imPOrtance relatIons NeTwork (POINT) 
as a classification pipeline.
%as it is easy to be constituted and extended with new modules. 
\whlie{Our model processes an arbitrary number of detected people in parallel}
% runs in parallel 
(as opposed to sequential relation modeling \cite{VIP_solomon2015vip,PR_li2018personrank}) and is fully differentiable \whli{(as opposed to the previous relation models using customized features \cite{PR_li2018personrank})}.
% /***Jason: whether this is difference as compared to previous work? if so, this should be put in the related work but not here******weihong: it is the same as the Relation Network for Object Detection.***/. 
For the \whlie{$i^{th}$} person $\mathbf{p}_i$ in an image, its label (i.e., important or non-important person) $\mathbf{s}_i$ 
% /***Jason: not confirmed***/ 
is estimated by:
\begin{equation}
\small
	\mathbf{s}_i =f^O(\mathbf{I}; \mathbf{p}_i|\theta^{O}) \circ f^R(\mathbf{f}^O_1,..., \mathbf{f}^O_N, \mathbf{f}^O_{global}|\theta^{R})\circ f^S(\mathbf{f}^{I}_i|\theta^{S}),
\end{equation}
where \whlia{$\circ$ denotes module composition,} \whlieee{$\mathbf{f}^{I}_i$ is the importance feature of $\mathbf{p}_i$} and $f^S(\mathbf{f}^{I}_i|\theta^{S})$ is the importance classification module parameterized by $\theta^{S}$, \whliee{which} follows the relation module $f^R(\cdot)$ parameterized by the parameter groups $\theta^{R}$. \whlia{In addition, the feature representation module $f^O(\mathbf{I}; \mathbf{p}_i|\theta^{O})$ parameterized by $\theta^{O}$ is employed to extract the \whli{person feature} $\mathbf{f}^O_i$ of $\mathbf{p}_i$ and the global feature $\mathbf{f}^O_{global}$ of the whole image \whlieee{$\mathbf{I}$} \whlieee{and $\circ$ is the operator to connect three modules.}}

% In addition, $\mathbf{f}^O_i$ is the \whli{person feature} of person $\mathbf{p}_i$ obtained by 
% %feeding the interior patch, the exterior patch and the location information (i.e. the coordinate of person in image) into 
% the feature representation module $f^O(\mathbf{I}; \mathbf{p}_i|\theta^{O})$, which is parameterized by $\theta^{O}$, and $\mathbf{f}^O_{global}$ is the feature of the whole image. 

The relation module $f^R(\cdot)$ \whlie{exploits the input} features of persons $\{\mathbf{p}_i\}_{i=1}^{N}$ and the global features $\mathbf{f}^O_{global}$ to automatically construct interaction graphs and encode effective relation features. Similar to existing attention modules \whlieee{\cite{AttenIsAll_vaswani2017attention}} and relational modules \whlieee{\cite{Rela_hu2018relation,Rela_zambaldi2018relational}}, we adopt the residual connection to aggregate the \whli{person feature} and the relation feature, resulting in a final importance feature $\mathbf{f}^{I}_i$, which comprises individual information, the relation information from other persons and the event information in an image. The details of each module are described in Section \ref{sec:Feat}, Section \ref{sec:Relation} and Section \ref{sec:Class}.

\subsection{Feature Representation Module}\label{sec:Feat}
% \vspace{-0.1cm}
Since \whlie{feature representation}
% people detection
is the first step in important people detection, we require the feature representation module (Figure \ref{fig:Feat}) to be capable of extracting effective features from local to global information (i.e., the people's interior/individual information, the exterior/contextual information around the people and the global information illustrating the event cues).
%diverse types of information. We propose a feature representation module to from multiple level 
%individual, contextual and global
As with most vision works \cite{Mulbox_szegedy2014scalable,PR_li2018personrank,Key_ramanathan2015detecting,VIP_solomon2015vip}, it is natural to use the information inside the bounding box of the detected person, called the interior patch in this work, to represent the person's interior/individual feature.
The location is also an indispensable element of a person's individual feature for illustrating the importance of the person and the coordinate of the person in the image is included in our feature. The reason is, from the photographer's perspective, when the images of an event are captured, the photographer tends to place the important people in the center of the image, and the important people usually look clearer than other people in the image. 
\begin{figure}[t]
	\begin{center}
		\label{fig:Eq4more}
		%		\fbox{\rule{0pt}{2in}\rule{0.9\linewidth}{0pt}}
		\includegraphics[width=0.6\linewidth]{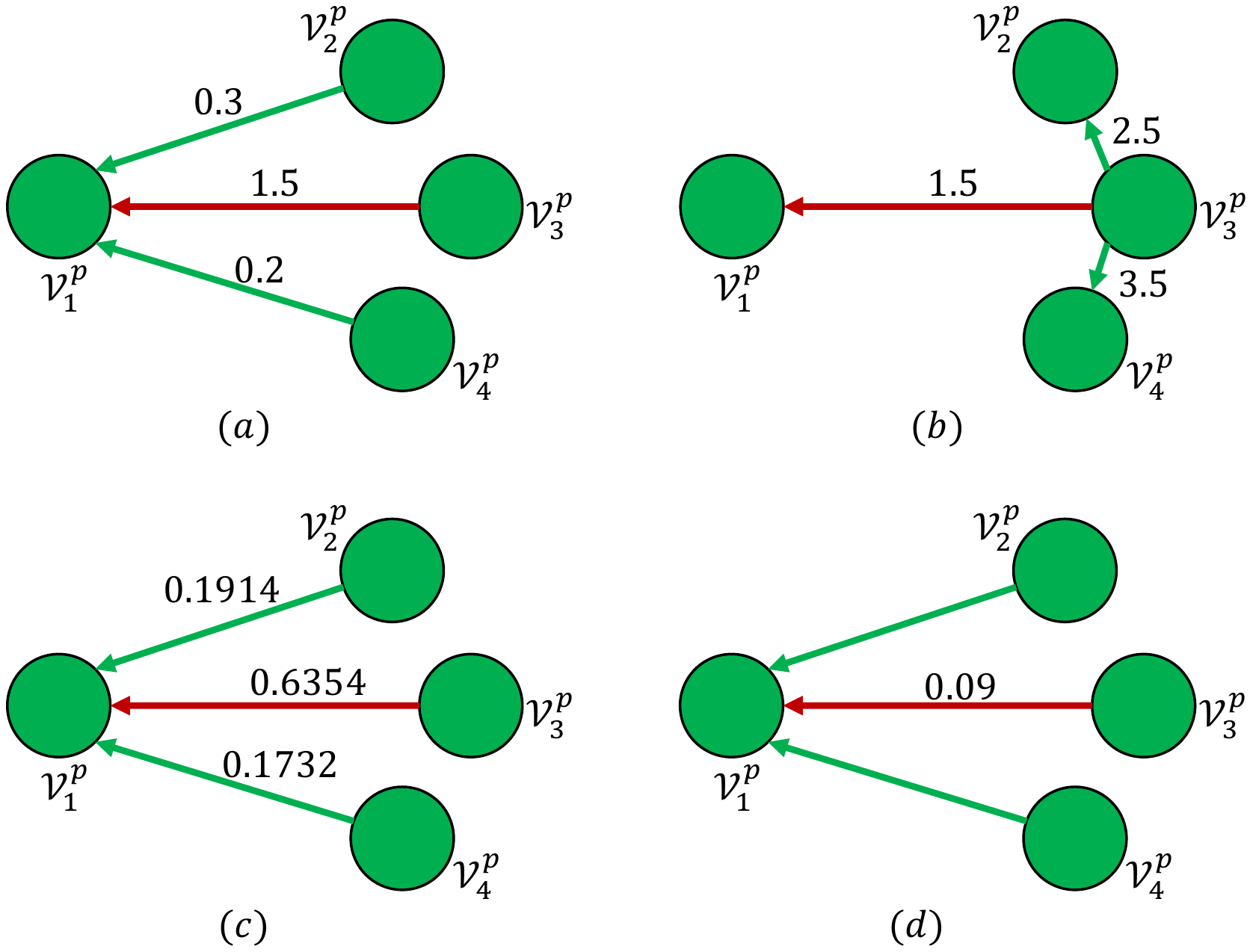}
		%		\vspace{-0.2cm}
		\centering\small\caption{Figure (a) and (b) present the input person-person interactions of $\mathcal{V}_1^p$ and the output person-person interactions of $\mathcal{V}_3^p$. Our method (i.e., Eq. (\ref{eq:softmax}) weakens the effect of the interaction from $\mathcal{V}_3^p$ to $\mathcal{V}_1^p$ \whlie{(the red link)} as $\mathcal{V}_3^p$ has too many outputs (Figure (d)). The attention model \cite{AttenIsAll_vaswani2017attention} treats each node equally, and the interaction from $\mathcal{V}_3^p$ to $\mathcal{V}_1^p$ has a larger impact (Figure (c)).}
		\label{fig:Eq4more}
	\end{center}
	% \vspace{-0.8cm}
\end{figure}
Additionally, the exterior/contextual information around each person must be considered for analyzing the importance of persons as this more global information, for instance, some objects that the person uses can aid in distinguishing the important people from the non-important people. 
%To enable our model to encode feature that embraces contextual feature, 
For this purpose, for each person, we crop an exterior patch\footnote{In this work, $C$ is trained on the validation set. Details of extracting the exterior patch and $C$ are reported in the Supplementary Material}, which is an image patch inside a box that is centered on the person's bounding box and is $C^2$ times larger than the scale of the person's bounding box.

In this work, 
\whliee{we use the ResNet-50 to extract features from} 
% we extract the ResNet-50 feature \cite{Img_he2015deep} for 
each interior and exterior patch 
because it has demonstrated its superiority in terms of important people detection \cite{PR_li2018personrank} and other vision tasks such as object detection \cite{SSD_liu2016ssd}.
As shown in Figure \ref{fig:Feat}, for each person in an image, we feed the interior and the exterior patches into separate Resnet-50s, transforming them into two $7\times 7\times 2048$ features (i.e., the interior feature and the exterior feature). While the coordinate is a four dimensional vector, we produce a heat map, which is a $224\times 224$ grid where one or several cells correspond to the person's coordinate are assigned as 1 and the others zero. We apply \whlie{convolutional kernels} to this heat map to produce a $7\times 7\times 256$ feature. Then, we concatenate the interior, the exterior and the location features, resulting in a $7\times 7\times 4352$ feature and employ two convolutional layers with one fully-connected (fc) layer to transform this concatenated feature into a $1024$ dimensional vector $\mathbf{f}^O_i$, called the \emph{person feature}. 
% /***Jason: i change it. please keep consistent in the whole paper******weihong: Done.***/ 

As the important person is inevitably related to the event that the person is involved in, 
the global information that represents this event should be considered as well. 
Similar to the interior and the exterior features, the whole image (denoted as the global patch) is fed into another deep network, which comprises the convolutional layers of the ResNet-50, two additional convolutional layers and one fc layer for encoding a $1024$ dimensional $\mathbf{f}^O_{global}$. We call this feature the \emph{global feature}.

\subsection{Relation Module}\label{sec:Relation}
% \vspace{-0.1cm}
Given the person feature and the global feature, we aim to design a relation module that can encode the effective importance feature by aggregating the relation feature and the person feature. 
%Inspired by the success of attention module \cite{AttenIsAll_vaswani2017attention}, 
More specifically, \whli{we aggregate $r$ relation features encoded by $r$ parallel \whlie{\textbf{relation submodules}}\footnote{The structure of these relation submodules are the same while the parameters are NOT shared, which \whlcam{enables} POINT to automatically learn various types of relations.} and concatenate them into one relation feature vector. Then, we employ the residual connection to merge the relation feature and the person feature, yielding the \whlie{\emph{importance feature}} for each person $\mathbf{p}_i$}:
% and define the importance feature of $\mathbf{p}_i$ as:
\begin{equation}\label{eq:rela1}
\small
\mathbf{f}^{I}_i = \mathbf{f}^{O}_i + Concat[\mathbf{f}^{{R}^1}_i, \cdots, \mathbf{f}^{{R}^r}_i], (i=1,\cdots N),
\end{equation}
where $\mathbf{f}^{{R}^1}_i$ is a relation feature of person $\mathbf{p}_i$ computed by the first \whlie{relation submodule}. We use this parallel structure because it \whliee{it allows POINT to automatically model} various types of people relations and has been shown to be more effective in our work and others \cite{Rela_hu2018relation,AttenIsAll_vaswani2017attention}

\begin{figure}[t]
	\begin{center}
		\label{fig:Eq4}
%				\fbox{\rule{0pt}{2in}\rule{0.9\linewidth}{0pt}}
		\includegraphics[width=0.85\linewidth]{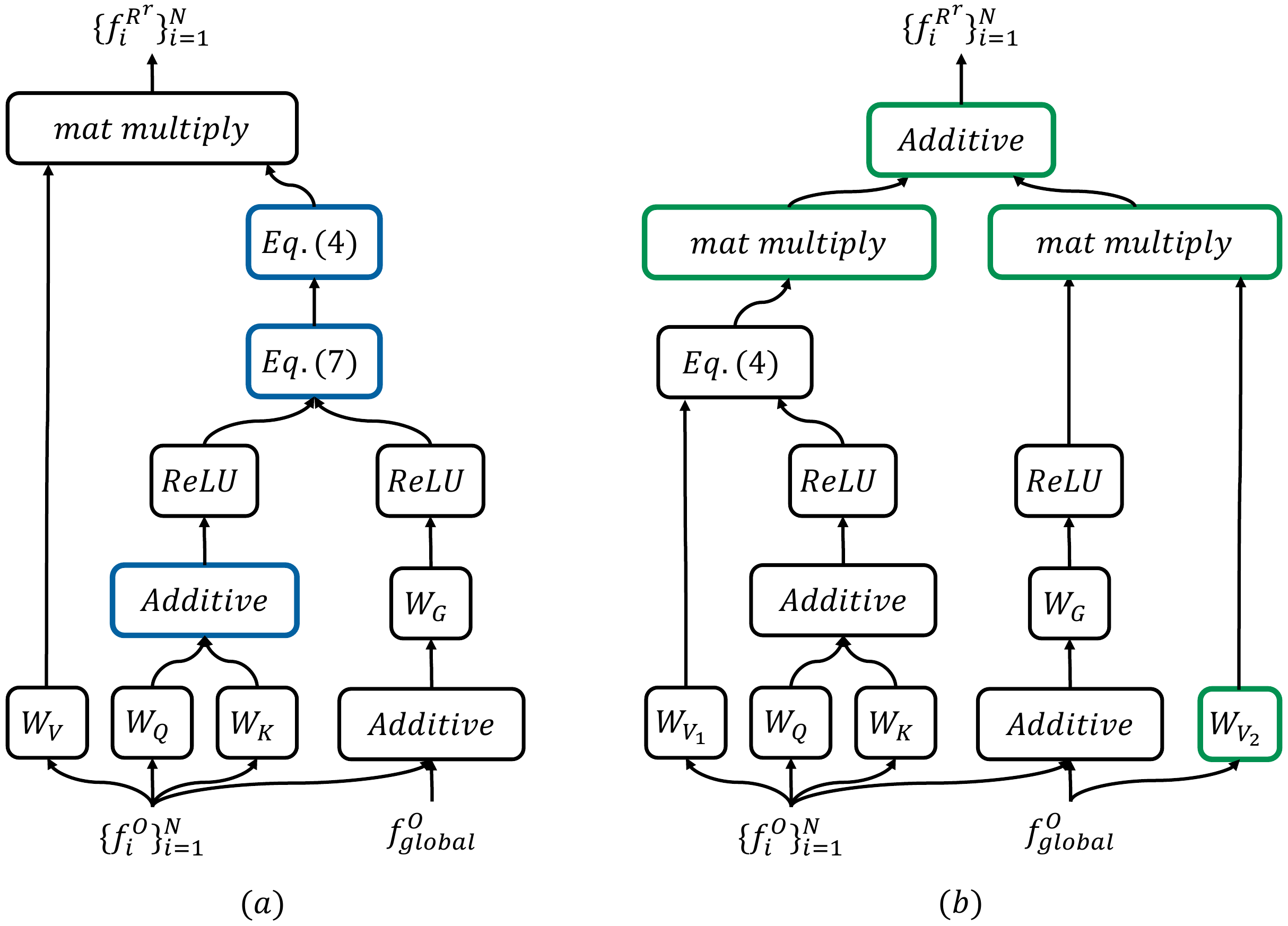}
		%		\vspace{-0.2cm}
		\centering\small\caption{Figure (a) and (b) illustrate two methods introduced in this work to embed global information into the \whlie{person-person interaction} (they are the illustrations of the relation submodule as well). Figure (a) is the method using Eq. (\ref{eq:relation1}) and Eq. (\ref{eq:global1}) while Figure (b) is the method using Eq. (\ref{eq:rela2}). The blue rectangle boxes show the difference between our method and the attention model \cite{AttenIsAll_vaswani2017attention} \whlie{and the relation module in \cite{Rela_hu2018relation,Rela_zambaldi2018relational}} while the green boxes illustrate the difference between the two methods we proposed. (Better viewed in color).}
%         /***Jason: make the border of the color bounding box thick ***/}
		\label{fig:Eq4}
	\end{center}
	\vspace{-0.35cm}
\end{figure}

\vspace{0.05cm}

\noindent \textbf{\whlie{Relations Modeling in the Relation Submodule.}}
We now describe \whlie{our importance relation computation} \whlie{in the \whlie{$\ell^{th}$ $(\ell=1,...,r)$ relation submodule}}. For each given image with $N$ detected persons, we obtain a feature set $\{\mathbf{f}^O_1,..., \mathbf{f}^O_N, \mathbf{f}^O_{global}\}$, and then the relation feature $\mathbf{f}^{{R}}_i$ with respect to the $i^{th}$ person is computed by
\begin{equation}\label{eq:relation1}
\small
\mathbf{f}^{R}_i = \sum_{j=1}^{N}\mathcal{E}_{ji} \cdot (\mathbf{W}_{V}\mathbf{f}^{O}_{j}).
\end{equation}
% \label{eq:relation1}
Here, we remove the superscript of \whlie{$\mathbf{f}^{{R}^\ell}_i$} and use $\mathbf{f}^{R}_i$ for description convenience.
The output of Eq. (\ref{eq:relation1}) aggregates the feature from the others by a weighted sum of the person features from the other people, and is linearly transformed by $\mathbf{W}_{V}$. 
%This structure is similar to the basic attention module, called “Scaled Dot-Product Attention” in \cite{AttenIsAll_vaswani2017attention}./***写这么多这些你方法别人觉得就没创新了***/
We formulate \whlie{$\mathcal{E}_{ji}$}, the \whlie{\emph{importance relation}} indicating the impact from the other people by:
\begin{equation}\label{eq:softmax}
\small
\mathcal{E}_{ji} = \frac{exp(\hat{\mathcal{E}}^{p}_{ji})}{\sum_{k=1}^{N}exp(\hat{\mathcal{E}}^{p}_{jk})},
\end{equation}
where $\hat{\mathcal{E}}^{p}_{ji}$ is the \whlie{\emph{importance interaction}} among persons \jason{and introduced in the following, and} it is estimated from both the person-person interaction graph and the event-person interaction graph.
Here, we compute the importance relation from person $\mathbf{p}_j$ to person $\mathbf{p}_i$ as the importance interaction from person $\mathbf{p}_j$ to person $\mathbf{p}_i$ scaled by the summation of the output \whlie{importance} interactions of person $\mathbf{p}_j$. 
%This is  and is shown to be convincing. 
Inspired by the PageRank algorithm \cite{PR_leskovec2014mining}, our model reflects the fact that an \whliee{importance interaction} from a node that has too many \whliee{importance interaction outputs} is less important, and this weakens the effect of the \whliee{importance interaction} on the importance relation (Figure \ref{fig:Eq4more}).

\vspace{0.05cm}

\noindent \textbf{\whli{Constructing Interaction Graphs.}}
% We now describe the importance interaction \whlie{$\hat{\mathcal{E}}^{p}_{ji}$} computation estimated from both interaction graphs. W
\whlie{In order to estimate the importance interaction $\hat{\mathcal{E}}^{p}_{ji}$,}
we first create the \whlie{\emph{person-person interaction graph}} and \whlie{\emph{event-person interaction graph}}, which are defined as $\mathcal{H}^{p}=(\mathcal{V}^{p}, \mathcal{E}^{p})$ and $\mathcal{H}^{g}=(\mathcal{V}^{g}, \mathcal{E}^{g})$, respectively. Here, $\mathcal{V}^{p}=\{\mathcal{V}_i^{p}\}_{i=1}^{N}$ are nodes representing persons and $\mathcal{V}^{g} = \{\mathcal{V}_i^{p}\}_{i=1}^{N} \cup \{\mathcal{V}^{e}\}$ are nodes in $\mathcal{H}^{g}$, where $\mathcal{V}^{e}$ is a node representing the event occurring in the image.
In addition, each element $\mathcal{E}^{p}_{ji}$ in $\mathcal{E}^{p}$ models the person-person interaction from $\mathbf{p}_j$ to $\mathbf{p}_i$ indicating \whlie{how $\mathbf{p}_j$ is interacting with \whlia{$\mathbf{p}_i$}, }
% the importance similarity between $\mathbf{p}_j$ and $\mathbf{p}_i$, 
and each element $\mathcal{E}^{g}_i$ \whliee{in $\mathcal{E}^{g}$} represents the event-person interaction indicating the probability of a person \whlie{being} involved in the event.
% being important.

In the person-person interaction graph $\mathcal{H}^{p}$, the interaction between pairwise persons is computed by \whlie{the \textbf{person-person interaction module}, which is} an additive attention function \cite{AttenFun_britz2017massive,AttenFunc_bahdanau2014neural}\footnote{\whlcam{
% the importance similarity between both persons. 
There are two commonly used attention functions/mechanisms: the additive attention function \cite{AttenFunc_bahdanau2014neural} and the less expensive scale dot product function \cite{AttenDot_luong2015effective,AttenIsAll_vaswani2017attention}. While the two are similar in theoretical complexity, the additive operation slightly and consistently outperforms the scale dot product operation \cite{AttenFun_britz2017massive}. This outcome is also verified in our experiments, so we use the additive attention function for the person-person interaction modeling.}}:
\begin{equation}\label{eq:add}
\small
\mathcal{E}^{p}_{ji} =  \whlcam{max\{0, \mathbf{w}_{P} \cdot (\mathbf{W}_{Q}\mathbf{f}^{O}_{i}+\mathbf{W}_{K}\mathbf{f}^{O}_{j}) \},}
\end{equation}
where both $\mathbf{W}_{Q}$ and $\mathbf{W}_{K}$ are matrices that project the \whli{person features} $\mathbf{f}^{O}_{i}$ and $\mathbf{f}^{O}_{j}$ into subspaces \whlcam{and the vector $\mathbf{w}_{P}$ is applied to measure how $\mathbf{p}_j$ is interacting with $\mathbf{p}_i$ in the subspace. Additionally, the $\text{max}\{\cdot \}$ function is employed to trim the person-person interaction at zero if the person is not interacting with the other person.}

% \whlie{and the $\text{max}\{\cdot \}$ function is employed to activate the effective feature in the subspace for \whliee{interaction modeling}. \whliee{Additionally}, the \whliee{vector} 
% % /***Jason: vector or matrix? small case for vector?***/ 
% \whlieee{$\mathbf{w}_{P}$} is applied to measure \whlie{how $\mathbf{p}_j$ is interacting with \whlia{$\mathbf{p}_i$}}.}
% the importance similarity between both persons. 
% In fact, there are two commonly used attention functions/mechanisms: the additive attention function \cite{AttenFunc_bahdanau2014neural} and the less expensive scale dot product function \cite{AttenDot_luong2015effective,AttenIsAll_vaswani2017attention}. While the two are similar in theoretical complexity, the additive operation slightly and consistently outperforms the scale dot product operation \cite{AttenFun_britz2017massive}. This outcome is also verified in our experiments, so we use the additive attention function for the person-person interaction modeling.

In the meantime, we estimate the event-person interaction by \whlie{the \textbf{event-person interaction module
\footnote{\whlcam{Eq. \ref{eq:global} is different from Eq. \ref{eq:add} as the event-person interaction differs from the person-person interaction (asymmetric) presenting how a person is interacting with another people: the event-person interaction should be equal to the person-event interaction (symmetric) and is estimated to find whether a person is involved in the event. 
}}
}}:
\begin{equation}\label{eq:global}
\small
\mathcal{E}^{g}_i = max\{0, \whlieee{\mathbf{w}_{G}}\cdot (\mathbf{f}^{O}_{i}+\mathbf{f}^{O}_{global})\},
\end{equation}
\whli{where $\mathbf{f}^{O}_{i}+\mathbf{f}^{O}_{global}$ is transformed into a scalar weight by \whlieee{$\mathbf{w}_{G}$} to indicate the probability of the person ($\mathbf{p}_i$) \whlie{being involved in the event.}}
% being important.}
\whlie{The event-person interaction is trimmed at 0, acting as a ReLU nonlinearity. The zero trimming operation restricts the event-person interactions only of the people being not related to the event.}
% person/***Jason: people or person ***/ not related /***Jason: not related or related ***/ to the event.}

% On both interaction functions Eq. (\ref{eq:add}) and Eq. (\ref{eq:global}), the interaction is trimmed at 0, acting as a ReLU nonlinearity. The zero trimming operation restricts the person-person interactions only between persons \whlie{without interaction (i.e., both persons do not interact with each other),}
% % a certain importance relationship  
% and it restricts the event-person interactions only of the person not related to the event.
% /***Jason: it is unclear for event-person interactions***//***weihong: I have add a sentence to explain it. I think it is clear now?***/.

\begin{figure}[t]
	\begin{center}
		\label{fig:Eq78}
%		\fbox{\rule{0pt}{2in}\rule{0.9\linewidth}{0pt}}
				\includegraphics[width=0.55\linewidth]{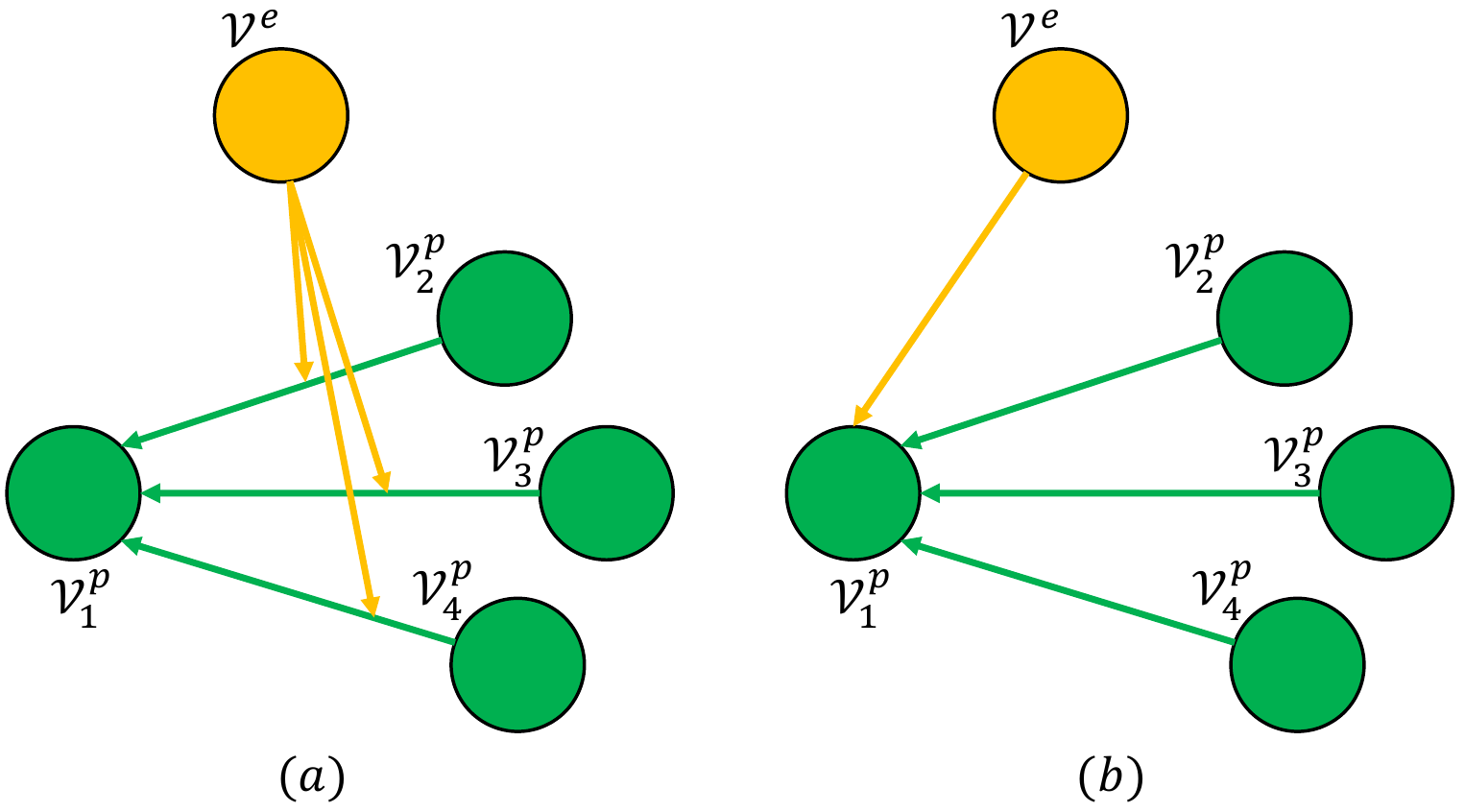}
		%		\vspace{-0.2cm}
		\centering\small\caption{We introduce two methods to \whlie{integrate the event-person interaction graph with the person-person interaction graph.}
%         integrate extra information into the \whlie{importance relation}.
        First, we treat the event-person interaction as the prior importance acting as the regulator to adjust the weight of the person-person interactions (Figure (a)). Second, we treat the event-person interaction as an extra input link for each person (Figure (b)).}
		\label{fig:Eq78}
	\end{center}
	\vspace{-0.35cm}
\end{figure}

\vspace{0.05cm}

\noindent \textbf{\whlie{Estimating Importance Interaction from Both Graphs.}}
Since we have two interaction graphs, the person-person interaction graph and the event-person interaction graph, the method for estimating the importance interaction \whlie{$\hat{\mathcal{E}}^{p}_{ji}$} from both graphs can significantly affect the importance relation computation and then impact the final results. In this work, we introduce two methods (Figure \ref{fig:Eq78}) to estimate the importance interaction from multiple graphs. Intuitively, we treat the event-person \whlie{interaction} as a prior importance and estimate the importance interaction \whlie{$\hat{\mathcal{E}}^{p}_{ji}$} as:
\begin{equation}\label{eq:global1}
\small
\hat{\mathcal{E}}^{p}_{ji} =  \mathcal{E}^{p}_{ji}\cdot \mathcal{E}^{g}_{j}.
\end{equation}
The advantage of this strategy is that the prior importance $\mathcal{E}^{g}_{j}$ acts as a regulator to adjust the effect of the \whlie{person-person} interaction $\mathcal{E}^{p}_{ji}$ on aggregating the relation features by enhancing the effect when the prior importance is large and reducing the impact in the opposite case.

An alternative strategy is to treat the event-person interaction as an additional graph to the \whlie{person-person interaction graph}. In other words, \whliee{we define the importance interaction as the person-person interaction} (i.e., $\hat{\mathcal{E}}^{p}_{ji} = \mathcal{E}^{p}_{ji}$) and the relation feature is aggregated as:
\begin{equation}\label{eq:rela2}
\small
\mathbf{f}^{R}_{i} = \sum_{j=1}^{N}\mathcal{E}_{ji} \cdot (\mathbf{W}_{V_1}\mathbf{f}^{O}_{j}) + \mathcal{E}^{g}_{i} \cdot(\mathbf{W}_{V_2}\mathbf{f}_{global}),
\end{equation}
where $\mathcal{E}_{ji}$ is computed by Eq. (\ref{eq:softmax}).
% where $\mathcal{E}_{ji}=\frac{exp(\mathcal{E}^{p}_{ji})}{\sum_{k=1}^{N}exp(\mathcal{E}^{p}_{jk})}$ \whlie{is the importance interaction}.
Here, the relation feature aggregates the feature from the others by a weighted sum of \whli{person features}
% original features
from the other people, linearly transformed by $\mathbf{W}_{V_1}$ and the global feature transformed by $\mathbf{W}_{V_2}$. In this way, the global information can be considered \whlie{during} encoding the importance features without affecting the effect of the person-person interaction. 

The above two strategies are verified to be effective for combining both person-person interactions
% among persons 
and event-person interactions, and they have comparable results.

% \noindent \textbf{Summary/***Jason: seems i am not right for this title. Please correct. Also, this subsection is too long, please add more sub-titles to help the reader to follow***/}. 
\vspace{0.05cm}
\noindent \textbf{\whli{Parameters of the Relation Module.}}
The \whlie{relation module} Eq. (\ref{eq:rela1}) is summarized in Figure \ref{fig:FrameWork}. It is easy to implement using basic operators, as illustrated in Figure \ref{fig:Eq4}. As the dimension of the output feature is the same as the input feature, 
\whliee{we can stack more than one relation module ($N_r$ relation modules) to refine the importance feature.}
% we can treat the relation module, which contains $r$ \whlie{relation submodules} as a \textbf{relation module block} and stack more than one relation module blocks ($N_r$ relation module blocks) to refine the importance feature. 
\whlieee{In Eq. (\ref{eq:rela1}), since we have $r$ \whlie{relation submodules} in one relation module, the parameters are $5\times r$ projections: $ \theta^{R}=\{\mathbf{W}^\ell_Q\in\mathcal{R}^{d_f\times d_k}, \mathbf{W}^{\ell}_K\in\mathcal{R}^{d_f\times d_k}, \mathbf{W}^{\ell}_V\in\mathcal{R}^{d_f\times d_v}, \mathbf{w}^{\ell}_P\in\mathcal{R}^{d_f}, \mathbf{w}^{\ell}_G\in\mathcal{R}^{d_f}\}_{\ell=1}^{r}$, where $d_f=1024$ is the dimension of the \whli{person feature} and $d_k=d_v=\frac{d_f}{r}$}
% \whlie{$\mathbf{W}^\ell_Q\in\mathcal{R}^{d_f\times d_k}, \mathbf{W}^{\ell}_K\in\mathcal{R}^{d_f\times d_k}$, $ \mathbf{W}^{\ell}_V\in\mathcal{R}^{d_f\times d_v}$, $ \mathbf{W}^{\ell}_P\in\mathcal{R}^{d_f\times 1}$ and $\mathbf{W}^{\ell}_G\in\mathcal{R}^{d_f\times 1}$,} where $\ell=1,...,r$, $d_f=1024$ is the dimension of the \whli{person feature} and $d_k=d_v=\frac{d_f}{r}$. }
% More specifically, we employ $N_r=2$ relation module blocks which has $r=4$ parallel basic relation modules. For each parameter, we use $d_k=d_v=\frac{d_f}{r}=512$. 
Due to the reduced dimension of each relation submodule, the total computational cost is similar to that of the single relation submodule with full dimensionality.

\begin{table*}[tbp!]
	%	\vspace{-0.2cm}
	\centering
% 	\footnotesize
	\tiny
	\caption{\whlie{The mAP (\%) of Different Methods on both Datasets}}
	%	\vspace{-0.2cm}
	\resizebox{!}{0.8cm}
	{
% 		\scriptsize
		\small
		\begin{tabular}{c|c|c|c|c|c|c|c|c|c|c}
			\hline
            \hline
			\multirow{2}[2]{*}{Method} & Max-  & Max-  & Max-  & Most- & Max-  & SVR-  &\multirow{2}[2]{*}{VIP} & Ramanathan's & \multirow{2}[2]{*}{PR}&\textbf{Ours} \\
			& Face  & Pedestrian & Saliency & Center & Scale & Person &   &  model \cite{Key_ramanathan2015detecting}  & &($\textbf{POINT}$)  \\
			\hline
			\hline
			MS Dataset & 35.7 & 30.7 & 40.3 & 50.9 & 73.9 & 75.9 & {76.1} & -\ -& {\color{blue}{88.6}} & {\bf \color{red}{92.0}}\\
            \hline
            \hline
            NCAA Dataset   & 31.4 & 24.7 & 26.4 & 30.0 & 31.8 & {64.5} & 53.2 & 61.8 & {\color{blue}{74.1}} & {\bf \color{red}{97.3}}\\
			\hline
            \hline
		\end{tabular}%
	}
	\label{tab:MS_Methods}%
		% \vspace{-0.4cm}
\end{table*}%

\subsection{\whli{Classification Module} \jason{for End-to-End Learning}}\label{sec:Class}
% \vspace{-0.1cm}
After we obtain the importance feature for each person in an image, we utilize two fully connected layers (i.e., the classification module $f^S(\mathbf{f}^{I}_i|\theta^{S})$) to transform the feature into two scalar values indicating the probability of the person belonging to the \whlie{important people or non-important people classes}.
During training, the commonly used cross-entropy loss is employed to penalize the model, and the SGD is used to optimize the model \jason{for backward computation}. 
%Editor: Abbreviations and acronyms are typically defined the first time the term is used within the main text and then used throughout the remainder of the manuscript. Please consider adhering to this convention. The target journal may have a list of abbreviations that are considered common enough that they do not need to be defined.
During testing, the probability of \whlie{the important people class is used as the importance point for each people. In each image, and the people with the highest importance point will be selected as the most important people.}

% \vspace{-0.15cm}
\section{Experiments}
% \vspace{-0.15cm}
\whli{In this section, we conducted extensive experiments on two publicly available image-based important people detection datasets. We followed the standard evaluation protocol in the dataset \cite{PR_li2018personrank}. The mean average precision (mAP) and some visual comparisons are reported. The CMC curve, other visual results and the classification accuracy of all people tested are reported and analyzed in the Supplementary Material.} 
% More details about training and preprocessing are presented in the Supplementary Material.
% /***Jason: you can put more details about the training and processing here. We need some sample images in the paper, this is a vision conference***//***weihong: I have add the training details in this section. But I don't think we should put it in the paper. Details of the preprocessing is too long to put here.***/

% \vspace{-0.1cm}
\subsection{\todo{Datasets}}
% \vspace{-0.15cm}
For evaluation on important people detection in still images, there are two publicly available datasets \cite{PR_li2018personrank}: 1) The Multi-scene Important People Image Dataset (MS Dataset) and 2) the NCAA Basketball Image Dataset (NCAA Dataset).

\vspace{0.05cm}
\noindent \textbf{1) The MS Dataset}.
The MS Dataset contains 2310 images from more than six types of scenes. This dataset includes three subsets: a training set (924 images), a validation set (232 images), and a testing set (1154 images). The detected face bounding box and importance labels are provided.

% The MS Dataset contains 2310 images from more than six types of scene. This dataset includes three subsets: a training set consists of 924 images, a validation set consists of 232 images, and  a testing set consists of 1154 images. The detected bounding box (i.e. bounding box of face) and importance label of detected people are provided in this dataset.

\vspace{0.05cm}

\noindent \textbf{2) The NCAA Dataset}.
The NCAA Dataset is formed by extracting 9,736 frames of an event detection video dataset \cite{Key_ramanathan2015detecting} covering 10 different types of events. The person bounding box and the importance annotations are provided as well.

\renewcommand*{\arraystretch}{1.3}
\begin{table}[tpb]
	\centering
	\footnotesize
	\caption{\footnotesize \whlie{The mAP (\%) for Evaluating Different Components of our POINT on Both Datasets.}}
	% 	\vspace{-0.2cm}
	\resizebox{!}{1.42cm}
	{
		\scriptsize
		\begin{tabular}{c|c|c|c|c}
			%			\hline
			%			\multicolumn{4}{c||}{Multi-scene Important People Image Dataset} & \multicolumn{4}{c}{NCAA Basketball Image Dataset} \\
			\hline
            \hline
			Dataset &Method & mAP & Method & mAP \\
			\hline
			\multirow{3}[3]{*}{MS Dataset} &$\text{Base}^{\text{Inter}}$ & 72.6 &$\text{POINT}^{\text{Inter}}$ &\textbf{76.5} \\
			%			\hline
			&$\text{Base}^{\text{Inter+Loca}}$ & 79.5&$\text{POINT}^{\text{Inter+Loca}}$ &\textbf{85.6} \\
			%			\hline
%			&$\text{Base}^{\text{Indi+Cont}}$ & &$\text{POINT}^{\text{Indi+Cont}}$ & \\
			%			\hline
			&$\text{Base}^{\text{Inter+Exter+Loca}}$ & \color{blue}{\textbf{89.2}} &$\text{POINT}^{\text{Inter+Exter+Loca}}$ &\color{red}{\textbf{92.0}}\\
			%			\hline
			%			&   &   &   &   &   &  &   &   &   &   &   &   &    \\
			\hline
			\multirow{3}[3]{*}{NCAA Dataset} & $\text{Base}^{\text{Inter}}$ & 89.1 & $\text{POINT}^{\text{Inter}}$ & \textbf{90.3}\\
			%			\hline
			& $\text{Base}^{\text{Inter+Loca}}$ & 89.9 & $\text{POINT}^{\text{Inter+Loca}}$ &\textbf{93.9} \\
			%			\hline
%			& $\text{Base}^{\text{Indi+Cont}}$ & & $\text{POINT}^{\text{Indi+Cont}}$ & \\
			%			\hline
			& $\text{Base}^{\text{Inter+Exter+Loca}}$ & \color{blue}{\textbf{95.8}}& $\text{POINT}^{\text{Inter+Exter+Loca}}$ & \color{red}{{\textbf{97.3}}}\\
			\hline
            \hline
		\end{tabular}%
	}
	% \vspace{-0.2cm}
    \label{tab:Components}%
\end{table}%

% \vspace{-0.1cm}
\renewcommand*{\arraystretch}{1.3}
\begin{table}[tpb]
	\centering
	\footnotesize
	\caption{\footnotesize The mAP (\%) for Evaluating our Methods of Integrating Global Information on both Datasets.}
	% 	\vspace{-0.2cm}
	\resizebox{!}{1.25cm}
	{
		\scriptsize
		\begin{tabular}{c|c||c|c}
			\hline
            \hline
			\multicolumn{2}{c||}{MS Dataset} & \multicolumn{2}{c}{NCAA Dataset} \\
			\hline
			Method & mAP & Method & mAP\\
			\hline
			$\text{POINT}^{\mathcal{H}^{p}}$ &91.2 &  $\text{POINT}^{\mathcal{H}^{p}}$ &96.0 \\
			\hline
			$\text{POINT}^{\text{Eq. (\ref{eq:rela2})}}$ & 91.3&$\text{POINT}^{\text{Eq. (\ref{eq:rela2})}}$ & 96.7\\
			\hline
			$\text{POINT}^{\text{Eq. (\ref{eq:relation1})+Eq. (\ref{eq:global1})}}$ &\textbf{92.0} &  $\text{POINT}^{\text{Eq. (\ref{eq:relation1})+Eq. (\ref{eq:global1})}}$ &\textbf{97.3} \\
			
			%			\hline
			%			&   &   &   &   &   &  &   &   &   &   &   &   &    \\
			\hline
            \hline
		\end{tabular}%
	}
    % \vspace{-0.4cm}
	% 	\vspace{-0.5cm}
	\label{tab:Global}%
\end{table}%

% \vspace{-0.3cm}
\subsection{\todo{Comparison with Other Methods}}
% \vspace{-0.2cm}
We first compared our method with existing important people detection models: 1) the VIP model \cite{VIP_solomon2015vip}, 2) Ramanathan's model \cite{Key_ramanathan2015detecting} and 3) the PersonRank (PR) model \cite{PR_li2018personrank} as well as all baselines (i.e., max-face, max-pedestrian, max-saliency, most-center, max-scale and SVR-person) provided in \cite{PR_li2018personrank}. The experimental results are shown in Table \ref{tab:MS_Methods}\footnote{On the MS Dataset, we did not compare Ramanathan's model \cite{Key_ramanathan2015detecting} as it uses temporal information, which is not provided in the MS Dataset. All the results of other methods are from \cite{PR_li2018personrank}}. From the table, it is clear that our POINT obtains state-of-the-art results. It is noteworthy that our POINT achieves a significant improvement of 23.2 \% on the NCAA Dataset over the PersonRank method that achieved the best performance previously (i.e., 74.1 \%).
This verifies the efficacy of our POINT method for extracting higher level semantic feature that embraces more effective information for important people detection, compared to those customized or deep features trained for other tasks. \jason{This also indicates the effectiveness of incorporate the relation modeling with feature learning for important people detection.} 
%In addition to this, it is shown that 
Interestingly, the improvement on the MS Dataset is significantly less than that on the NCAA Dataset (i.e., 3.4\% vs 23.2\%, respectively). The reason is that there are limited numbers of images (i.e., 2310 images in total), which limited the training of our deep model, even though the data augmentation of the training data (such as RandomCrop) has been used on the MS Dataset.

%\noindent \textbf{Comparing Our Method for Estimating Importance Relation Weight with the Original Attention Weight in /weihong: cite a paper here/.}

\renewcommand*{\arraystretch}{1.3}
\begin{table}[tpb]
	\centering
	\footnotesize
	\caption{\footnotesize \whlie{The mAP (\%) for Comparison of our Method and the one in \cite{AttenIsAll_vaswani2017attention} for Estimating the \whlie{Importance Relation}  on both Datasets.}}
	% 	\vspace{-0.2cm}
	\resizebox{!}{1.0cm}
	{
		\scriptsize
		\begin{tabular}{c|c||c|c}
			\hline
            \hline
			\multicolumn{2}{c||}{MS Dataset} & \multicolumn{2}{c}{NCAA Dataset} \\
			\hline
			Method & mAP & Method & mAP\\
			\hline
			Attention \cite{AttenIsAll_vaswani2017attention} &90.0 &  Attention \cite{AttenIsAll_vaswani2017attention} &95.8 \\
			\hline
			Ours (POINT) & \textbf{92.0}&  Ours (POINT) &\textbf{97.3} \\
			%			\hline
			%			&   &   &   &   &   &  &   &   &   &   &   &   &    \\
			\hline
            \hline
		\end{tabular}%
	}
    % \vspace{-0.2cm}
	% 	\vspace{-0.5cm}
	\label{tab:Weight}%
\end{table}%

%\vspace{-0.2cm}
%\noindent \textbf{Baseline}. We compared our model with several baselines: ``Most-Center'', ``Max-Scale'', ``Max-Face'', ``Max-Pedestrian'' \cite{PD_nam2014local} and ``SVR-Person", where ``Most-Center'' means selecting the person who is closest to the center of an image, ``Max-Scale'' means selecting the person of whom the face/body size is the largest one in an image, ``Max-Face'' means selecting the person of whom the detected face is most confident in an image by using \cite{FaceDetector}, ``Max-Pedestrian'' means selecting the person of whom the pedestrian detection score is the most confident using \cite{PD_nam2014local}, and ``SVR-Person" means using $\nu\text{-SVR}$ \cite{SVM_tool,VIP_solomon2015vip} to predict relative importance between persons and select the most important person based on the concatenatation of the four types of features we used.

%	\vspace{0.1cm}

\renewcommand*{\arraystretch}{1.3}
\begin{table}[tpb]
	\centering
	\footnotesize
	\caption{\footnotesize The mAP (\%) for Evaluating the Effect of $r$ on Both Datasets}
	% 	\vspace{-0.2cm}
	\resizebox{!}{0.86cm}
	{
		\scriptsize
		\begin{tabular}{c|c|cccccc}
			\hline
            \hline
			 \multirow{2}[2]{*}{Dataset} & \multirow{2}[2]{*}{Baseline} & \multicolumn{6}{c}{Ours (POINT)}  \\
%              \hline
% 			\hline
			\cline{3-8} & & $r$=1 & $r$=2 & $r$=4  & $r$=8 & $r$=16  & $r$=32   \\
			\hline
			MS Dataset & 89.2 &  90.7 & 91.4  & \textbf{92.0}  & 91.4  & 91.8  & 91.4   \\
            \hline
%             \multirow{3}[3]{*}{NCAA Dataset} & Baseline & \multicolumn{6}{c}{Ours (POINT)} \\
%             \hline
%             &   & $r$=1  & $r$=2 & $r$=4 & $r$=8 & $r$=16 &$r$=32 \\
%             \hline
            NCAA Dataset & 95.8  &  96.2 &  96.8 & \textbf{97.3} & 96.8  &  97.0 &  96.6 \\
            \hline
            \hline
		\end{tabular}%
	}
	% \vspace{-0.2cm}
	\label{tab:valueofr}%
\end{table}%

\renewcommand*{\arraystretch}{1.3}
\begin{table}[tpb]
	\centering
	\footnotesize
	\caption{\footnotesize The mAP (\%) for Evaluating the Effect of $N_r$ on Both Datasets}
	% 	\vspace{-0.2cm}
	\resizebox{!}{0.95cm}
	{
		\scriptsize
		\begin{tabular}{c|c|ccccc}
			\hline
            \hline
			\multirow{2}{*}{Dataset} & \multirow{2}{*}{Baseline} & \multicolumn{4}{c}{Ours (POINT)}  \\
			\cline{3-6} & & $N_r$=1 & $N_r$=2 & $N_r$=4  & $N_r$=6    \\
            \hline
			MS Dataset & 89.18 &  91.96 & \textbf{91.97}  & 90.99  &  90.90     \\
			\hline
%             \hline
%             \multirow{3}[3]{*}{NCAA Dataset} & Baseline & \multicolumn{4}{c}{Ours (POINT)} \\
%             & & $N_r$=1  & $N_r$=2 & $N_r$=4 & $N_r$=6  \\
            NCAA Dataset &  95.84  &  97.28 & 97.24  & \textbf{97.29}  &  96.02 \\
            \hline
            \hline
		\end{tabular}%
	}
	% \vspace{-0.2cm}
	\label{tab:valueofNr}%
\end{table}%

\renewcommand*{\arraystretch}{1.3}
\begin{table}[tpb]
	\centering
	\footnotesize
	\caption{\footnotesize The mAP (\%) for Evaluating Different Types of Attention Functions on both Datasets.}
	% 	\vspace{-0.2cm}
	\resizebox{!}{1.0cm}
	{
		\scriptsize
		\begin{tabular}{c|c||c|c}
			\hline
            \hline
			\multicolumn{2}{c||}{MS Dataset} & \multicolumn{2}{c}{NCAA Dataset} \\
			\hline
			Method & mAP & Method & mAP\\
			\hline
			$\text{POINT}^\text{Scaled Dot Product}$ & 90.7&  $\text{POINT}^\text{Scaled Dot Product}$ &96.2 \\
            \hline
			$\text{POINT}^\text{Additive}$ & \textbf{92.0} &  $\text{POINT}^\text{Additive}$ &\textbf{97.3} \\
			%			\hline
			%			Element-wise Dot Product & &  Element-wise Dot Product & \\
			%			\hline
			%			&   &   &   &   &   &  &   &   &   &   &   &   &    \\
			\hline
            \hline
		\end{tabular}%
	}
	% \vspace{-0.4cm}
	\label{tab:attenfun}%
\end{table}%

\subsection{Evaluation of Our POINT}
% \vspace{-0.1cm}

\noindent \textbf{Evaluating Different Components of POINT.}
%From Section \ref{sec:Feat}, since we have 
Since there is a lack of end-to-end trainable deep learning models for important people detection, we form a baseline that only comprises the feature representation module and the importance classification module. This approach predicts the importance of persons without considering their relations with others and the event-person relations. It is defined as:
\begin{equation}
\small
\mathbf{s}^{Baseline}_i =f^O(\mathbf{p}_i|\theta^{O})
\circ f^S(\mathbf{f}^{O}_i|\theta^{S}).
\end{equation}
\whlie{It is formed to evaluate the effect of the relation module (i.e., our POINT)
% from the interior patch (i.e. the people's detected bounding box), the location feature from the coordinate of persons and the exterior/contextual feature from the exterior patch). We evaluated the effect of the relation module (i.e. our POINT)
% /***Jason: are the feature independent of POINT. We claim it is end-to-end, is it? If not, don't write end-to-end******weihong: Yes, our method is an end-to-end trainable model. the feature is trained for estimate relations***/ 
and different components of the feature (i.e., the interior feature, the location feature and the exterior/contextual feature).} 
% In addition to this, the proposed POINT uses features that consist of multiple cues (i.e., the interior feature, the location feature and the exterior/contextual feature). 
The results are reported in Table \ref{tab:Components} where the $\text{Base}^{\text{Inter}}$ indicates the baseline using only the interior feature and $\text{POINT}^{\text{Inter + Loca +Exter}}$ is our full model. The POINT, using the feature comprising all features, is described in Section \ref{sec:Feat}.
% the interior feature, the location feature and the exterior feature described in Section \ref{sec:Feat}.

From Table \ref{tab:Components}, it is noteworthy that our POINT consistently obtains better mAP values than the baseline using different types of features (e.g., 92.0\% vs 89.2\%, respectively, on the MS Dataset using three types of cues). This result indicates that embedding the relation module \whliee{introduced in this paper} can significantly aid in extracting more discriminant, higher level semantic information, which dramatically increases the performance. Additionally, we can see that both the baseline and POINT improve the mAP on important people detection by using more cues compared to those using less information or a single type of information (e.g., the $\text{Base}^{\text{Indi+Cont+Loca}}$ has an improvement of 16.6\% mAP over the $\text{Base}^{\text{Indi}}$, which obtains 72.6\% mAP on the MS Dataset). 
\begin{figure}[t]
	\begin{center}
		\label{fig:VisualComTF}
% 		\fbox{\rule{0pt}{2in}\rule{0.9\linewidth}{0pt}}
		\includegraphics[width=0.7\linewidth]{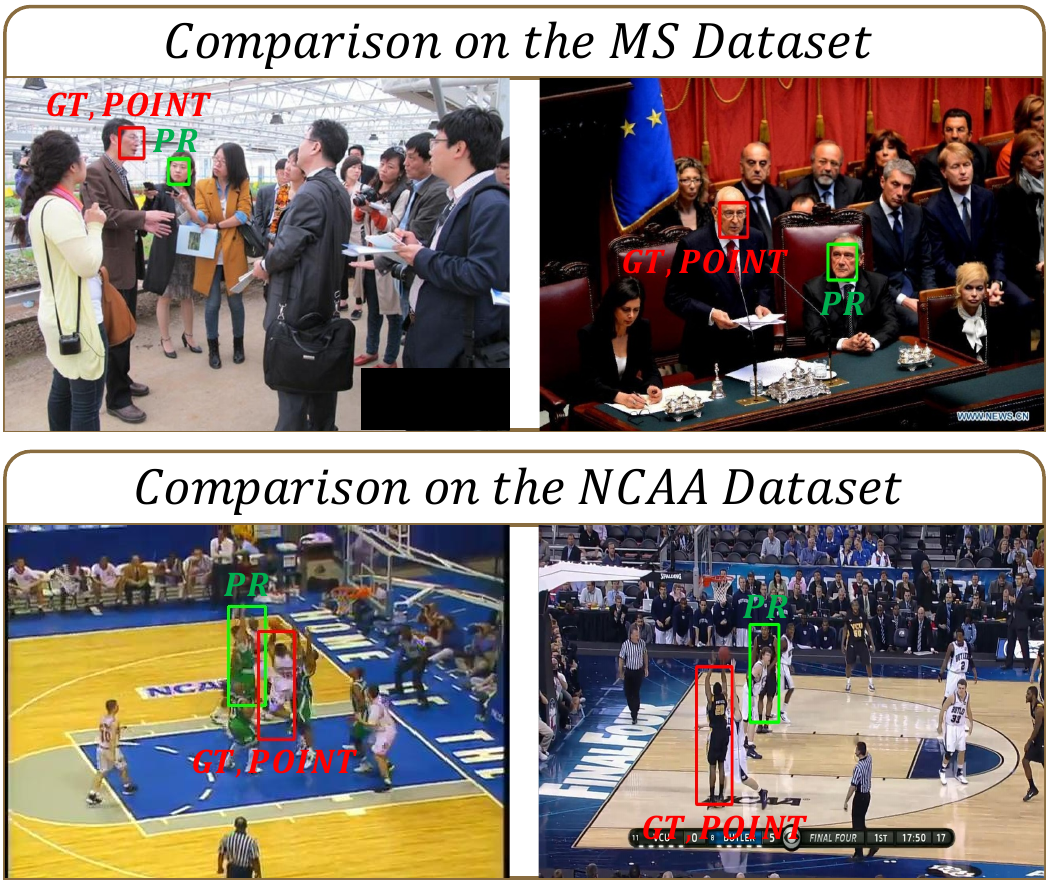}
		%		\vspace{-0.2cm}
		\centering\small\caption{Visual results of detecting important people and \jason{comparison} with related work (i.e., PersonRank (PR)) on Both Datasets.}
		\label{fig:VisualComTF}
	\end{center}
		\vspace{-0.35cm}

\end{figure}

\vspace{0.05cm}
\noindent \textbf{Integrating the Additional Global Information and Estimating the \whlie{Importance Relation}.}
In this work, we introduce two methods to integrate the event-person interaction graph with the \whlie{person-person interaction graph}. Table \ref{tab:Global} presents the results of our POINT for detecting important people without global information (i.e., $\text{POINT}^{\mathcal{H}^{p}}$), our POINT using the global information in different ways (i.e., $\text{POINT}^{\text{Eq. (\ref{eq:relation1})+Eq. (\ref{eq:global1}})}$ and $\text{POINT}^{\text{Eq. (\ref{eq:rela2}})}$). It is clearly shown that both methods successfully integrate the global information into the importance feature and improve the performance. In general, the improvement when using the global information as a prior importance is higher than that of treating the event-interaction graph as an additional graph (e.g., 1.3\% vs 0.7\%, respectively, on the NCAA Dataset). 

We also compared our method \whlie{of} estimating the \whlie{importance relation} with that of the attention weight \cite{AttenIsAll_vaswani2017attention} \whlie{(i.e., the relation module in other vision tasks \cite{Rela_hu2018relation,Rela_zambaldi2018relational}}), and the results on both datasets are reported in Table \ref{tab:Weight}. While the whole relation network is completely different from \cite{AttenIsAll_vaswani2017attention,Rela_hu2018relation,Rela_zambaldi2018relational} due to different tasks, it is clear that our \whlie{relation module} is more effective than the \whlie{relation model} used in \cite{AttenIsAll_vaswani2017attention,Rela_hu2018relation,Rela_zambaldi2018relational}, as we have consistent improvement (e.g., 92.0\% vs 90.0\%, respectively, on the MS Dataset). This result verifies the efficacy of Eq. (\ref{eq:softmax}).
% /***Jason: please check***/.

\vspace{0.05cm}

% \vspace{0.05cm}

\noindent \textbf{Visual Results and Comparisons.}
\whlie{In this section, selected visual results and comparisons are reported in Figure \ref{fig:VisualComTF} to further evaluate our POINT. 
% The results are shown in Figure \ref{fig:VisualComTF}.
As shown in Figure \ref{fig:VisualComTF}, it is clear that our POINT can detect the important people in some complex cases (e.g. in the both image in the second row, the defender and the shooter are very closed and our POINT can correctly assign most points to the shooter while the PersonRank (PR) usually pick the defender or other player as the important people.}

% \noindent \textbf{Methods for Embedding Location Information.}
% \whli{
% % In this work, we encode the location information in a different way as both individual location feature and  relation feature based on location information are important for important people detection, which is verified by the results in Table \ref{tab:Components}. 
% Both the individual location feature and the location relation feature are useful for important people detection, which is verified by the results in Table \ref{tab:Components}. 
% We further evaluate our location embedding method with another approach in \cite{Rela_hu2018relation}.
% The results on both datasets are shown in Table \ref{tab:LocationEmbedding}. It is clearly shown that our approach for embedding location information performs better than the location embedding method in \cite{Rela_hu2018relation} for important people detection.}

\vspace{0.05cm}
\noindent \textbf{Effect of $r$ and $N_r$ on Important People Detection.}
The number of \whlie{relation submodule} $r$ and the number of \whliee{stacked relation module} $N_r$ can slightly affect our POINT. To evaluate the effect of both parameters, we report the results of our POINT using $r$ ranging from 1 to 32 and keeping $N_r=1$ in Table \ref{tab:valueofr}. Then, we select $r=4$ as it yields the best result and set $N_r$ ranging from 1 to 6. \whli{The evaluation results of the effect of $N_r$ are reported in Table \ref{tab:valueofNr}. The results shows that using the $r>1$  \whliee{relation submodule} in a relation module enables our POINT to obtain better results because using multiple \whlie{relation submodules} allows our POINT to model various types of relations. In addition, we find that when we set $N_r>1$, the POINT obtains slightly better results (e.g., setting $N_r=2$ on the MS dataset and $N_r=4$ on the NCAA dataset are the best) because the added relation modules can aid in refining the importance features.}
% In fact, stacking $N_r>1$ relation module block can aids refining importance features. Interestingly, We find when we set $N_r>1$, the POINT can obtain slightly better results (e.g. setting $N_r=2$ on MS dataset and $N_r=4$ on NCAA dataset are the best).}

% The results in both tables show that using $r>1$ and $N_r>1$ enables our POINT to obtain better results as using multiple basic relation modules and relation module blocks allow our POINT to model various types of relation and refine the importance feature $N_r>1$ times. In addition to this, we find $r$ and $N_r$ can be set to appropriately small (i.e. $r=4$ and $N_r=2$) and good results can be achieved. 
%we find when $r$ and $N_r$ set to be larger than $2$, they are less sensitive to the results. Thus we can achieve a good results when selecting an appropriately small $r$ and $N_r$. 
% This is useful as it needs less computation cost.

\vspace{0.05cm}
\noindent \textbf{Evaluation of the Attention Functions.}
Currently, there are two commonly used attention functions for modeling interaction between any pairs of entities, the additive and the scaled dot product attention functions. Similar to \cite{AttenFun_britz2017massive}, we find the additive attention function works slightly but consistently better than the scaled dot product function \whlieee{from Table \ref{tab:attenfun}} (e.g., 97.3\% vs 96.2\%, respectively, on the NCAA Dataset).
\vspace{0.05cm}

\noindent \textbf{Running time.}
We implement our model using PyTorch on a machine with CPU E5 2686 2.3 GHz, GTX 1080 Ti and 256 GB RAM. The running time of our POINT for processing an image is sensitive to the number of persons in the image.
On average, POINT can process 10 frames per second (fps), which is significantly faster than the PersonRank (0.2 fps) and the VIP (0.06 fps). This result indicates that our POINT largely improves the speed of the important people detection model.

\section{Conclusion}
% \vspace{-0.2cm}

We have proposed a deep importance relation network to investigate deep learning for exploring and encoding the relation features and exploiting them for important people detection. 
%In the experiments, \whlie{the additive function performs the best on pair-wise relation modeling and four types of information (i.e., the interior, the location, the exterior and the global information) proposed in this work are shown to be effective for important people detection.
% we investigated and discussed the effect of various types of basic relation functions (i.e., the additive function and scaled dot product function) on modeling pairwise persons’ interactions and the effect of different types of information on important people detection. 
\jason{More importantly, we have shown that POINT successfully integrate the relation modeling with feature learning to learn the feature for relation modeling. In addition, POINT can learn to encode and exploit the relation feature for important people detection.
% our proposed POINT can be trained to automatically encode relation features without any additional supervision. 
It was clearly shown that our proposed POINT could obtain state-of-the-art performance on two public datasets.} 
% \vspace{-0.15cm}
\section{Acknowledgement}
% \vspace{-0.2cm}
This work was supported partially by the National Key Research and Development Program of China (2018YFB1004903), NSFC(61522115), and Guangdong Province Science and Technology Innovation Leading Talents (2016TX03X157).

{\small
\bibliographystyle{ieee}
\bibliography{POINT}

\begin{thebibliography}{10}\itemsep=-1pt

\bibitem{AttenFunc_bahdanau2014neural}
Dzmitry Bahdanau, Kyunghyun Cho, and Yoshua Bengio.
\newblock Neural machine translation by jointly learning to align and
  translate.
\newblock {\em International Conference on Machine Learning}, 2014.

\bibitem{IP_berg2012understanding}
Alexander~C Berg, Tamara~L Berg, Hal Daume, Jesse Dodge, Amit Goyal, Xufeng
  Han, Alyssa Mensch, Margaret Mitchell, Aneesh Sood, Karl Stratos, et~al.
\newblock Understanding and predicting importance in images.
\newblock In {\em Computer Vision and Pattern Recognition}, 2012.

\bibitem{AttenFun_britz2017massive}
Denny Britz, Anna Goldie, Minh-Thang Luong, and Quoc Le.
\newblock Massive exploration of neural machine translation architectures.
\newblock In {\em Conference on Empirical Methods in Natural Language
  Processing}, 2017.

\bibitem{Rela_hu2018relation}
Han Hu, Jiayuan Gu, Zheng Zhang, Jifeng Dai, and Yichen Wei.
\newblock Relation networks for object detection.
\newblock In {\em Computer Vision and Pattern Recognition}, 2018.

\bibitem{IP_hwang2012learning}
Sung~Ju Hwang and Kristen Grauman.
\newblock Learning the relative importance of objects from tagged images for
  retrieval and cross-modal search.
\newblock {\em International journal of computer vision}, 100(2):134--153,
  2012.

\bibitem{IP_le2007finding}
Duy-Dinh Le, Shin'ichi Satoh, Michael~E Houle, and Dat Phuoc~Tat Nguyen.
\newblock Finding important people in large news video databases using
  multimodal and clustering analysis.
\newblock In {\em International Conference on Data Engineering}, 2007.

\bibitem{IP_lee2012discovering}
Yong~Jae Lee, Joydeep Ghosh, and Kristen Grauman.
\newblock Discovering important people and objects for egocentric video
  summarization.
\newblock In {\em Computer Vision and Pattern Recognition}, 2012.

\bibitem{IP_lee2015predicting}
Yong~Jae Lee and Kristen Grauman.
\newblock Predicting important objects for egocentric video summarization.
\newblock {\em International Journal of Computer Vision}, 114(1):38--55, 2015.

\bibitem{PR_leskovec2014mining}
Jure Leskovec, Anand Rajaraman, and Jeffrey~David Ullman.
\newblock {\em Mining of massive datasets}.
\newblock Cambridge University Press, 2014.

\bibitem{PR_li2018personrank}
Wei-Hong Li, Benchao Li, and Wei-Shi Zheng.
\newblock Personrank: Detecting important people in images.
\newblock In {\em International Conference on Automatic Face \& Gesture
  Recognition}, 2018.

\bibitem{SSD_liu2016ssd}
Wei Liu, Dragomir Anguelov, Dumitru Erhan, Christian Szegedy, Scott Reed,
  Cheng-Yang Fu, and Alexander~C Berg.
\newblock Ssd: Single shot multibox detector.
\newblock In {\em European conference on computer vision}, 2016.

\bibitem{AttenDot_luong2015effective}
Minh-Thang Luong, Hieu Pham, and Christopher~D Manning.
\newblock Effective approaches to attention-based neural machine translation.
\newblock {\em Conference on Empirical Methods in Natural Language Processing},
  2015.

\bibitem{Key_ramanathan2015detecting}
Vignesh Ramanathan, Jonathan Huang, Sami Abu-El-Haija, Alexander Gorban, Kevin
  Murphy, and Li Fei-Fei.
\newblock Detecting events and key actors in multi-person videos.
\newblock {\em Computer Vision and Pattern Recognition}, 2016.

\bibitem{VIP_solomon2015vip}
Clint Solomon~Mathialagan, Andrew~C Gallagher, and Dhruv Batra.
\newblock Vip: Finding important people in images.
\newblock In {\em Computer Vision and Pattern Recognition}, 2015.

\bibitem{IP_spain2011measuring}
Merrielle Spain and Pietro Perona.
\newblock Measuring and predicting object importance.
\newblock {\em International Journal of Computer Vision}, 91(1):59--76, 2011.

\bibitem{Mulbox_szegedy2014scalable}
Christian Szegedy, Scott Reed, Dumitru Erhan, and Dragomir Anguelov.
\newblock Scalable, high-quality object detection.
\newblock {\em arXiv}, 2014.

\bibitem{Acti_tang2017latent}
Yongyi Tang, Peizhen Zhang, Jian-Fang Hu, and Wei-Shi Zheng.
\newblock Latent embeddings for collective activity recognition.
\newblock In {\em Advanced Video and Signal Based Surveillance}, 2017.

\bibitem{AttenIsAll_vaswani2017attention}
Ashish Vaswani, Noam Shazeer, Niki Parmar, Jakob Uszkoreit, Llion Jones,
  Aidan~N Gomez, {\L}ukasz Kaiser, and Illia Polosukhin.
\newblock Attention is all you need.
\newblock In {\em Advances in Neural Information Processing Systems}, 2017.

\bibitem{RN_wang2017appearance}
Limin Wang, Wei Li, Wen Li, and Luc Van~Gool.
\newblock Appearance-and-relation networks for video classification.
\newblock In {\em Computer Vision and Pattern Recognition}, 2018.

\bibitem{Rela_xu2015show}
Kelvin Xu, Jimmy Ba, Ryan Kiros, Kyunghyun Cho, Aaron Courville, Ruslan
  Salakhudinov, Rich Zemel, and Yoshua Bengio.
\newblock Show, attend and tell: Neural image caption generation with visual
  attention.
\newblock In {\em International conference on machine learning}, 2015.

\bibitem{RN_yang2018learning}
Flood Sung~Yongxin Yang, Li Zhang, Tao Xiang, Philip~HS Torr, and Timothy~M
  Hospedales.
\newblock Learning to compare: Relation network for few-shot learning.
\newblock In {\em Computer Vision and Pattern Recognition}, 2018.

\bibitem{Rela_zambaldi2018relational}
Vinicius Zambaldi, David Raposo, Adam Santoro, Victor Bapst, Yujia Li, Igor
  Babuschkin, Karl Tuyls, David Reichert, Timothy Lillicrap, Edward Lockhart,
  et~al.
\newblock Relational deep reinforcement learning.
\newblock {\em arXiv}, 2018.

\end{thebibliography}
}

\end{document}